%% file: main.tex
\documentclass[11pt]{article}
\usepackage[T1]{fontenc}
\usepackage[left=1in, right=1in, top=1in]{geometry}
\usepackage{xargs}
\usepackage[numbers]{natbib}
\usepackage{fancyhdr}
\usepackage{setspace}
\usepackage{lastpage}
\usepackage{upgreek}
\usepackage{amsmath,mathtools,amsthm,amsfonts,amssymb}	
\usepackage{amssymb,dsfont,bbm}	
\usepackage{xcolor}  
\usepackage[textwidth=3cm,textsize=footnotesize]{todonotes}
\usepackage{bm}
\usepackage{enumitem}
\usepackage{mathrsfs}
\usepackage{wrapfig}
\usepackage{booktabs}
\usepackage{subcaption}
\usepackage{caption}
\usepackage{xspace}

\usepackage{listings}
\lstdefinestyle{gfnxpython}{
  language=Python,
  basicstyle=\ttfamily\small,
  keywordstyle=\color{blue},
  commentstyle=\color{gray},
  stringstyle=\color{teal},
  showstringspaces=false,
  frame=single,
  breaklines=true,
  columns=fullflexible
}

\setcitestyle{number}

\usepackage[colorlinks=true,breaklinks=true,bookmarks=true,urlcolor=blue,citecolor=blue,linkcolor=blue,bookmarksopen=false,draft=false]{hyperref}

\usepackage{aliascnt}
\usepackage{cleveref}

\usepackage{authblk}

\input{def}

\begin{document}

\title{\texttt{gfnx}: Fast and Scalable Library for \\ Generative Flow Networks in JAX}

\author[1]{D. Tiapkin}
\author[2]{A. Agarkov}
\author[2]{N. Morozov}
\author[2]{I. Maksimov}
\author[2]{\\A. Tsyganov}
\author[3,2]{T. Gritsaev}
\author[2]{S. Samsonov}

\affil[1]{École Polytechnique, Palaiseau, France}
\affil[2]{HSE University, Moscow, Russia}
\affil[3]{Constructor University, Bremen, Germany}

\renewcommand\thefootnote{\fnsymbol{footnote}} 
\footnotetext[1]{\texttt{daniil.tiapkin@polytechnique.edu, aagarkov@hse.ru, nvmorozov@hse.ru, yvmaksimov@hse.ru, atsyganov@hse.ru, tgritsaev@constructor.university, svsamsonov@hse.ru}}

\date{}

\maketitle

\begin{abstract}
In this paper, we present \texttt{gfnx}, a fast and scalable package for training and evaluating Generative Flow Networks (GFlowNets) written in JAX. \texttt{gfnx} provides an extensive set of environments and metrics for benchmarking, accompanied with single‑file implementations of core objectives for training GFlowNets. We include synthetic hypergrids, multiple sequence generation environments with various editing regimes and particular reward designs for molecular generation, phylogenetic tree construction, Bayesian structure learning, and sampling from the Ising model energy. Across different tasks, \texttt{gfnx} achieves significant wall‑clock speedups compared to Pytorch-based benchmarks (such as \texttt{torchgfn} library \cite{lahlou2023torchgfn}) and author implementations. For example, \texttt{gfnx} achieves up to $55$ times speedup on CPU-based sequence generation environments, and up to $80$ times speedup with the GPU-based Bayesian network structure learning setup. Our package provides a diverse set of benchmarks and aims to standardize empirical evaluation and accelerate research and applications of GFlowNets. The library is available on GitHub
(\url{https://github.com/d-tiapkin/gfnx}) and on pypi (\url{https://pypi.org/project/gfnx/}). Documentation is available on \url{https://gfnx.readthedocs.io}.
\end{abstract}

\input{src/intro}

\input{src/method}

\input{src/experiments}

\input{src/conclusion}

\bibliography{refs}

\appendix
\part*{Appendix}
\input{app/background}

\input{app/environment_details}

\input{app/experimental_details}

\end{document}

%% file: def.tex
\def\nset{\mathbb{N}}

\let\originalleft\left
\let\originalright\right
\renewcommand{\left}{\mathopen{}\mathclose\bgroup\originalleft}
\renewcommand{\right}{\aftergroup\egroup\originalright}

\renewcommand{\epsilon}{\varepsilon}

\newcommand{\cE}{\mathcal{E}}

\newcommand{\cL}{\mathcal{L}}

\newcommand{\cP}{\mathcal{P}}

\newcommand{\cR}{\mathcal{R}}

\newcommand{\cT}{\mathcal{T}}

\newcommand{\cX}{\mathcal{X}}

\renewcommand{\phi}{\varphi}

\def\eqsp{\,}

\renewcommand{\top}{\mathsf{\scriptscriptstyle T}}

\newcommand{\TB}{\texttt{TB}\xspace}
\newcommand{\SubTB}{\texttt{SubTB}\xspace}

\usepackage{xcolor}
\definecolor{Bleu}{RGB}{0,0,204}
\definecolor{Violet}{RGB}{102,0,204}
\definecolor{Rouge}{RGB}{204,0,0}
\definecolor{Highlight}{RGB}{251,0,0}
\definecolor{Gray}{gray}{0.85}
\definecolor{LightGray}{gray}{0.9}

\usepackage{xcolor}
\definecolor{mydarkblue}{rgb}{0,0.08,0.45}
\definecolor{dmorange500}{HTML}{FF5F19}
\definecolor{dmblue300}{HTML}{2267EB}
\definecolor{dmred300}{HTML}{FF617B}
\hypersetup{
	colorlinks,
	citecolor=dmblue300,
	linkcolor=dmred300,
	urlcolor=mydarkblue
}

\newcommand{\PF}{\cP_{\mathrm{F}}}
\newcommand{\PB}{\cP_{\mathrm{B}}}
\newcommand{\rmZ}{\mathrm{Z}}
\newcommand{\Ptraj}[1]{\mathsf{P}_{\cT}^{#1}}

%% file: src/intro.tex
\section{Introduction}
\label{sec:intro}
\begin{figure}[!ht]
    \centering
    \begin{subfigure}{0.5\linewidth}
        \centering
        \includegraphics[width=0.6\linewidth]{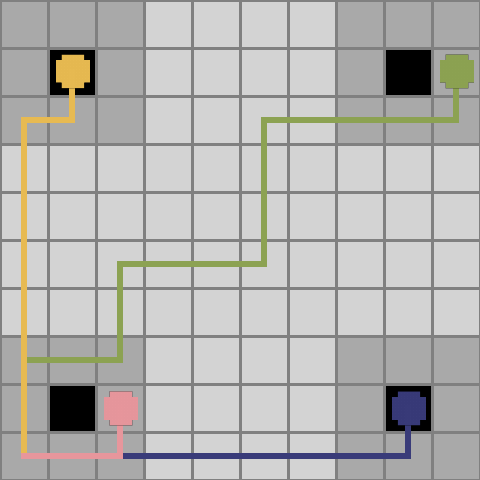}
        \subcaption{2D Hypergrid environment.}
    \end{subfigure}\hfill
    \begin{subfigure}{0.5\linewidth}
        \centering
        \includegraphics[width=0.535\linewidth]{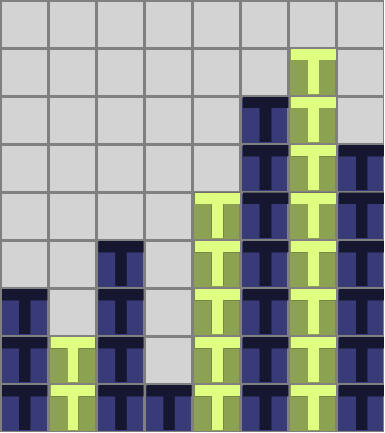}
        \subcaption{Bit Sequence environment.}
    \end{subfigure}
    \caption{Visualization of GFlowNet environments. The first figure illustrates four sample trajectories, each represented by distinct colored paths from initial to terminal states. The second figure illustrates the sequential construction of a bit string; each row corresponds to a different environment state: the first row represents the empty string (initial state), the last row the complete string (final state), and intermediate rows display the token-by-token generation process.}
    \label{fig:intro:hypergrid}
\end{figure}

Many problems in machine learning and scientific discovery require drawing samples from complex, discrete distributions, often known only up to a normalizing constant. The well-known solution to this problem is to use MCMC methods \cite{neal2011mcmc,robert1999monte}. These methods typically require many sampling steps to converge, making them prohibitively costly as the problem scale increases. The \textit{amortized sampling} approach provides an alternative: instead of running an MCMC sampler, one trains a generative model to perform \emph{approximate} sampling from the target distribution. A recent class of models addressing this problem is given by the \emph{Generative Flow Networks} (GFlowNets, GFNs, \cite{bengio2021flow}). GFlowNets have shown promising results across a broad range of domains, such as phylogenetic tree generation~\cite{zhou2024phylogfn}, crystal structure discovery~\cite{mistal2023crystalgfn,nguyen2023hierarchical}, biological sequence design~\cite{jain2022biological,yin2025synergy}, and molecular generation~\cite{koziarski2024rgfn,shen2024tacogfn}.
\par 
Mathematical foundations of GFlowNets were established in \cite{bengio2021flow,bengio2021gflownet}. Recent works establish connections between GFlowNets and (hierarchical) variational inference algorithms \cite{malkin2023gflownets}, and formalize their connections with reinforcement learning (RL) \cite{tiapkin2024generative,mohammadpour2024maximum,deleu2024discrete}, particularly with a class of RL problems with entropy regularization. Given these important contributions, the majority of recent papers have focused on applications of GFlowNets to particular problems, raising a need for a scalable and standardized benchmark package to evaluate new methods.

Several libraries provide GFlowNet training and environment frameworks, including \texttt{torchgfn}~\cite{lahlou2023torchgfn}, \texttt{recursionpharma/gflownet}~\cite{recursionpharama/gflownet}, and \texttt{alexhernandezgarcia/gflownet}~\cite{hernandez-garcia2024}. These toolkits implement essential functionalities and offer helpful abstractions, but they share a common limitation: environment logic typically executes on the host (CPU) and thus does not fully leverage specialized hardware acceleration. As a result, data must be repeatedly transferred between CPU and accelerator hardware (GPU or TPU), creating a performance bottleneck during training.

Recent trends in RL research highlight that runtime efficiency is critical for developing new algorithms. Fast environments allow rapid iteration, large-scale hyperparameter sweeps, and more reliable statistical comparisons across random seeds~\cite{lu2022discovered,jackson2024discovering,colas2018randomseedsstatisticalpower,jordan2024benchmarking,eimer2023hyperparameters}. A key technique enabling such acceleration is just-in-time (JIT) compilation of the entire training loop, achievable through JAX~\cite{jax2018github}. This approach has led to substantial speedups across RL systems~\cite{lu2022discovered,flair2024jaxmarl,bortkiewicz2025accelerating,evojax2022,coward2024JaxUED,nishimori2024jaxcorl}, supported by a vast ecosystem of JAX-native environments and tools~\cite{brax2021github,gymnax2022github,koyamada2023pgx,bonnet2024jumanji,matthews2024craftax,nikulin2023xlandminigrid,flashbax,deepmind2020jax}.

\paragraph{Our contribution.}
In this work, we introduce a JAX-based library for GFlowNets. Our goal is to bring JAX-compiled environments and training utilities to the GFlowNet community, enabling accelerated research and offering efficient experimental pipelines. The library also follows the "single-file" design philosophy of CleanRL~\cite{huang2022cleanrl}, providing a clear and reliable baseline implementation.

%% file: src/method.tex
\section{Library Details}
\label{sec:documentation}
The core of the \texttt{gfnx} library consists of JIT-able environments, reward modules, and success metrics implemented entirely in JAX \cite{jax2018github}. Each component is documented on \url{https://gfnx.readthedocs.io} with the corresponding source papers, usage examples, and explanatory notes.

For every environment, we provide a concise, single-file JAX baseline that mirrors the simplicity and hackability of \texttt{CleanRL} \cite{huang2022cleanrl}. These examples rely on \texttt{Equinox} \cite{kidger2021equinox} for neural networks and are JIT-compiled end-to-end together with the environment, following the practices popularized by \texttt{purejaxrl} \cite{lu2022discovered}.

\paragraph{Library architecture.} The main part of the library is organized into the following modules:
\begin{itemize}
    \item \texttt{base.py} — Defines the shared primitives (\texttt{BaseEnvState}, \texttt{BaseEnvParams}, \texttt{BaseVecEnvironment}, \texttt{BaseRewardModule}) and their typing aliases that appear throughout the codebase.
    \item \texttt{environment/} — Implements vectorized, JITable environments derived from \texttt{BaseVecEnvironment}, pairing an \texttt{EnvState} to hold mutable state with \texttt{EnvParams} that configure dynamics and bind reward modules.
    \item \texttt{reward/} — Implements reward modules and likelihood/prior components for each environment. By decoupling rewards from dynamics we support swapping reward families or learning them during GFlowNet training without recompiling environment logic.
    \item \texttt{metrics/} — Provides success metrics for monitoring sampling quality. Multiple metrics highlight how GFlowNet evaluation differs from standard RL, where raw return is the main score.
    \item \texttt{utils/} — Gathers helper routines ranging from environment-specific utilities to reusable tools such as \texttt{gfnx.utils.forward\_rollout}, which produces trajectories for any fixed environment.
\end{itemize}

In addition, the repository offers the following companion code outside the core package, intended as stable, hackable examples rather than prescriptive frameworks:
\begin{itemize}
    \item \texttt{baselines/} — Single-file, end-to-end JAX training scripts that demonstrate fast compiled loops and provide reproducible implementations of established algorithms.
    \item \texttt{proxy/} — Utilities for training dataset-driven proxy reward models; these scripts tie into the AMP environment section for deeper discussion.
\end{itemize}

Together these components emphasize modularity, full JAX compatibility, and a hackable workflow that lowers the barrier to experimenting with new GFlowNet ideas.

\vspace{0.3cm}

We highlight a few implementation details:
\begin{itemize}
    \item Following the JAX paradigm, environments are \textit{stateless}; all mutable data lives in the \texttt{EnvState} object returned by \texttt{env.reset} and modified explicitly by each \texttt{env.step}.
    \item Environments are vectorized to simplify reward evaluation. We only evaluate rewards when at least one element of the batch is terminal by wrapping the computation in \texttt{jax.lax.cond}, which avoids the redundant work that a naive \texttt{jax.vmap} over terminal checks would incur.
    \item Instead of returning raw rewards, environments emit \texttt{log\_reward}: terminal transitions yield their log-reward, while non-terminal steps return zero. Log-scale rewards align with how most GFlowNet algorithms consume the signal.
\end{itemize}

\paragraph{Example.} 

A short example from the documentation illustrates how to instantiate the Hypergrid environment, roll it forward, and use the stop action:

\vspace{-0.2cm}

\begin{lstlisting}[style=gfnxpython, caption={Minimal Hypergrid usage in \texttt{gfnx}.}]
import jax
import jax.numpy as jnp
import gfnx

# 1. Pick a reward module;
reward = gfnx.EasyHypergridRewardModule()

# 2. Create the environment. Need to pass a corresponding reward module
env = gfnx.HypergridEnvironment(reward_module=reward, dim=3, side=5)
params = env.init(jax.random.PRNGKey(0))

# 3. Reset to get the initial observation/state batch. 
obs, state = env.reset(num_envs=1, env_params=params)

# 4. Take a forward step (increment coordinate 0).
action = jnp.array([0], dtype=jnp.int32)
obs, state, log_reward, done, _ = env.step(state, action, params)

print("Terminal?", bool(state.is_terminal[0]))  # returns False
print("Reward (log scale):", float(log_reward[0]))  # return 0

# 5. Stop when you are ready to terminate the trajectory. Stop action is the last action.
stop = jnp.array([env.action_space.n - 1], dtype=jnp.int32)
obs, state, log_reward, done, _ = env.step(state, stop, params)

print("Terminal?", bool(state.is_terminal[0]))  # returns True
print("Reward (log scale):", float(log_reward[0]))  # returns a corresponding log-reward
\end{lstlisting}

\vspace{-0.35cm}

\paragraph{Comparison with \texttt{torchgfn}.} \texttt{torchgfn} ships with tightly integrated training and environment stacks, whereas \texttt{gfnx} deliberately decouples these pieces so researchers can mix and match environment logic, reward modules, and training code. This separation trades the ready-made training loops of \texttt{torchgfn} for greater flexibility when running fully on-device execution or embedding the environments inside custom trainers.

\vspace{-0.2cm}

\begin{lstlisting}[style=gfnxpython, caption={Example of backward transtions in \texttt{gfnx}.}]
# 1. Reset the environment
obs, state = env.reset(num_envs=1, env_params=params)

# 2. Take a forward step (increment coordinate 0).
action = jnp.array([0], dtype=jnp.int32)
next_obs, next_state, log_reward, done, _ = env.step(state, action, params)

# 3. Compute a corresponding backward action
bwd_action = env.get_backward_action(state, action, next_state, params)

# 4. Make a backward step
_, prev_next_state, _, _, _ = env.backward_step(next_state, bwd_action, params)

# 5. Verify that we indeed inverted the action
import chex; chex.assert_trees_all_equal(state, prev_next_state)     # Do not raise error
\end{lstlisting}

The libraries also diverge in how they treat backward actions: \texttt{torchgfn} defines a backward move as an inverse to every forward transition (e.g., "remove a particular symbol on a particular position" in sequence generation), while \texttt{gfnx} abstracts backward actions to the structural choices alone (e.g., "remove any character on a particular position"). This abstraction mirrors our broader design goal of keeping the environment design as flexible as possible. We intentionally design backward transitions to mirror their forward counterparts as closely as possible, which makes it easy to reason about state reversibility and implement symmetric training objectives. In particular, this symmetry allows us to implement a backward rollout by only replacing the initial states by terminal ones and \texttt{env.step} by \texttt{env.backward\_step}.

%% file: src/experiments.tex
\section{Environments and Runtime}
\label{sec:experiments}
\texttt{gfnx} implements a wide range of environments from GFlowNet literature:

\begin{enumerate}
    \item \textbf{Hypergrids \cite{bengio2021flow}.} This environment is a $d$-dimensional discrete hypercube of side length $H$. There are actions that increment a single coordinate by $1$ while remaining in the grid and a stop action. The reward concentrates in high-reward regions near the corners of the hypercube, with substantially lower rewards elsewhere.
    \item \textbf{Bit sequences \cite{tiapkin2024generative}.}
    In this environment our goal is to sample bit strings of a fixed length $n$ in non-autoregressive manner (See Appendix \ref{app:seq_exp} for a discussion on different generation schemes for sequences). The string is split into $k$-bit blocks, and an empty token is initially put into each position. The generation process replaces and empty token with a $k$-bit word at each step until no empty tokens remain in the sequence. The vocabulary is, therefore, predefined as well as the mode set which we use for evaluation setup. The reward is defined through Hamming distance to the closest mode. 
    \item \textbf{TFBind8 \cite{shen2023towards}.} TFBind8 environment evaluates DNA sequence design, where the goal is to generate sequences of nucleotides. The reward is wet-lab measured DNA binding activity to a human transcription factor, SIX6 \cite{barrera2016survey}. Formally, this is an autoregressive sequence generation environment of fixed length $8$ with a vocabulary of size $4$ (corresponding to nucleotides).
    \item \textbf{QM9 \cite{shen2023towards}.} QM9 is a small molecule generation environment. We adopt the prepend/append sequence formulation, which models the generation process using 11 building blocks with 2 stems, producing molecules of 5 blocks. Rewards are given by a proxy model trained to predict the HOMO-LUMO gap, an important molecular property \cite{zhang2020molecularmechanicsdrivengraphneural}.
    \item \textbf{AMP \cite{jain2022biological}.}
    In the AMP task, our objective is to generate peptides with antimicrobial properties. We model this as an autoregressive environment with variable-length sequences (up to 60 tokens) over a 20–amino acid vocabulary. The environment provides rewards from a proxy model trained on the DBAASP database~\cite{pirtskhalava2021dbaasp}. 
    \item \textbf{Phylogenetic trees \cite{zhou2024phylogfn}.} 
    The environment starts from a forest of $n$ singleton species and proceeds by repeatedly merging two subtrees into a new common ancestor. The trajectory has a fixed length of $n - 1$ steps, after which a rooted binary tree is formed. States correspond to intermediate forests and actions to pairwise merges. The terminal reward induces a Gibbs distribution that favors trees requiring fewer mutations. Only the tree topology is modeled, with branch lengths excluded.
    \item \textbf{Bayesian structure learning \citep{deleu2022bayesian}.} 
    The environment constructs a Directed Acyclic Graph by sequentially adding edges while enforcing acyclicity. Any state may be terminal, with the reward equal to the log-posterior under linear-Gaussian or BGe scores. Datasets are generated from an Erd\H{o}s--R\'enyi ground-truth graph using a linear-Gaussian model.
    \item \textbf{Ising model \citep{zhang2022generative}.} 
    The environment represents partial spin assignments of a lattice, with states given by ternary vectors $s \in \{-1, +1, \varnothing\}^{N \times N}$ and initial state containing only unassigned sites (i.e. $\varnothing$). At each step, an action chooses an unassigned site and sets its spin to $-1$ or $+1$, producing a full configuration after $N^2$ steps. The reward corresponds to the Gibbs distribution of an Ising model with energy function $\mathcal{E}_{J}(\mathbf{x}) = -\mathbf{x}^\top J \mathbf{x}$, where $J \in \mathbb{R}^{N^2 \times N^2}$, and $\mathbf{x} \in \mathbb{R}^{N^2}$.
\end{enumerate}

Runtime comparisons of our library with standard baselines are listed in Table \ref{tab:time_exps}. Some environments are sufficiently small to allow benchmarking on a CPU, while others are benchmarked on a GPU. The \texttt{gfnx} implementation achieves up to an 80-fold speedup on the Bayesian structure learning environment (GPU) and up to a 55-fold speedup on the QM9 environment (CPU). 

\begin{table}[!ht]
    \centering
    \begin{tabular}{l|l|l|l|ll}
        \toprule
        Environment & Baseline & Device & Objective & Baseline & \texttt{gfnx} \\
        & Impl. &&&\\
        \midrule
        Hypergrid ($20\!\times\!20\!\times\!20\!\times\!20$) & \cite{lahlou2023torchgfn} & CPU & \texttt{DB} & 178.3$_{\pm2.0}$ it/s & \textbf{1560.0$_{\pm3.6}$} it/s \\
        &&&\texttt{TB} & 219.9$_{\pm1.1}$ it/s & \textbf{1463.3$_{\pm1.4}$} it/s \\
        &&&\texttt{SubTB} & 121.0$_{\pm0.7}$ it/s & \textbf{596.0$_{\pm0.3}$} it/s \\
        \hline
        Bitseq ($n=120, k=8$) & \cite{tiapkin2024generative} & GPU & \texttt{DB} & 52.4$_{\pm0.8}$ it/s & \textbf{1666.4$_{\pm8.0}$} it/s \\
        &&&\texttt{TB} & 54.0$_{\pm0.2}$ it/s & \textbf{2433.6$_{\pm67.1}$} it/s \\
        \hline
        TFBind8 & \cite{shen2023towards} & CPU & \texttt{TB} & 230.4$_{\pm2.7}$ it/s & \textbf{6929.3$_{\pm26.0}$} it/s \\
        \hline
        QM9 & \cite{shen2023towards} & CPU & \texttt{TB} & 162.3$_{\pm1.5}$ it/s & \textbf{9061.6$_{\pm28.7}$} it/s \\
        \hline
        AMP & \cite{jain2022biological} & GPU & \texttt{TB} & 21.2$_{\pm1.7}$ it/s & \textbf{413.1$_{\pm3.4}$} it/s \\
        \hline
        Phylogenetic Trees (DS-1) & \cite{deleu2024discrete} & GPU & \texttt{FLDB} & 13.3$_{\pm0.7}$ it/s & \textbf{263.6$_{\pm2.1}$} it/s \\
        Phylogenetic Trees (DS-2) & & & & 9.9$_{\pm0.1}$ it/s & \textbf{208.4$_{\pm4.0}$} it/s \\
        Phylogenetic Trees (DS-3) & & & & 8.6$_{\pm0.1}$ it/s & \textbf{149.7$_{\pm2.4}$} it/s \\
        Phylogenetic Trees (DS-4) & & & & 10.0$_{\pm0.1}$ it/s & \textbf{122.0$_{\pm1.0}$} it/s \\
        Phylogenetic Trees (DS-5) & & & & 11.7$_{\pm0.7}$ it/s & \textbf{76.1$_{\pm0.9}$} it/s \\
        Phylogenetic Trees (DS-6) & & & & 6.3$_{\pm0.3}$ it/s & \textbf{66.3$_{\pm1.4}$} it/s \\
        Phylogenetic Trees (DS-7) & & & & 2.9$_{\pm0.1}$ it/s & \textbf{36.7$_{\pm0.4}$} it/s \\
        Phylogenetic Trees (DS-8) & & & & 4.1$_{\pm0.2}$ it/s & \textbf{39.1$_{\pm0.6}$} it/s \\
        \hline
        Structure Learning & \cite{lahlou2023torchgfn} & GPU & \texttt{MDB} & 0.73$_{\pm0.03}$ it/s & \textbf{58.0$_{\pm1.0}$} it/s \\
        \hline
        Ising model ($N=9$) & -- & GPU & \texttt{TB} & -- & \textbf{27.8$_{\pm0.2}$} it/s \\
        Ising model ($N=10$) & -- & GPU & \texttt{TB} & -- & \textbf{26.4$_{\pm0.6}$} it/s \\
        \bottomrule
    \end{tabular}
    \caption{Iterations per second comparison between baseline and \texttt{gfnx} implementations on GFlowNet environments. Higher values indicate better performance. We add the $3$ sigma standard error interval for time performance. For Ising model we implemented the EBM setting from~\cite{zhang2022generative}, which had no available open-source implementations.}
    \label{tab:time_exps}
\end{table}

Full comparisons with implementations from previous literature across various metrics, along with detailed experimental setups, are presented in \Cref{app:env}.

%% file: src/conclusion.tex
\section{Conclusion \& Future Work}

Overall, \texttt{gfnx} provides a suite of JAX-native environments, reward modules, and metrics that speeds up amortized-sampling research while preserving the transparency needed for reproducible benchmarks. The single-file, end-to-end compiled baselines run up to $80\times$ faster than comparable non-JAX implementations without degrading sampling quality, lowering the barrier to rapid experimentation. Despite this progress, several gaps remain and motivate further developments:
\begin{itemize}
    \item \textit{Continuous action spaces}: in the current form, \texttt{gfnx} supports only discrete action spaces, yet many problems, such as a more complete version of phylogenetic tree generation environment \cite{zhou2024phylogfn} also have continuous components;
    \item \textit{Non-acyclic environments}: many natural problems, such as permutation generation, might be more clearly written without an acyclicity constraint~\cite{brunswic2024theory,morozov2025revisiting};
    \item \textit{Multi-objective support}: the need to generate diverse Pareto optimal solutions in multi-objective scenarios also arises in many problems \cite{jain2023multi, chen2023order};
    \item \textit{Additional  baselines}: GFlowNets can be successfully trained with various entropy-regularized RL algorithms \cite{tiapkin2024generative, deleu2024discrete, mohammadpour2024maximum, morozov2024improving, he2025random}. It will also be valuable to implement backward policy optimization algorithms~\cite{jang2024pessimistic, gritsaev2025optimizing} and exploration techniques~\cite{kim2023local, madan2025towards, kim2025adaptive} that were shown to have various benefits for GFlowNet training;
    
    \item \textit{Trainer vectorization}: batching over seeds and hyperparameters, in the style of \texttt{purejaxrl} \cite{lu2022discovered}, would improve reproducibility and hyperparameter sweeps on small environments.
\end{itemize}

%% file: app/background.tex
\section{Background}
\label{sec:background}

\paragraph{Generative Flow Networks.} GFlowNets aim at sampling from a probability distribution over a finite discrete space \(\mathcal{X}\) given by $\pi(x) = \cR(x) / \mathrm{Z}$, where $\cR \colon \mathcal{X} \to \mathbb{R}_{\geq 0}$ is the \textit{GFlowNet reward}, and $\mathrm{Z}$ is the (typically unknown) normalizing constant.

To formally define a generation process in GFlowNets, we introduce a directed acyclic graph (DAG) \(\mathcal{G} = (\mathcal{S}, \mathcal{E})\), where \(\mathcal{S}\) is a state space and \(\mathcal{E} \subseteq \mathcal{S} \times \mathcal{S}\) is a set of edges (or transitions). There is exactly one state, $s_0$, with no incoming edges, which we refer to as the \textit{initial state}. All other states can be reached from $s_0$, and the set of \textit{terminal states} with no outgoing edges coincides with the space of interest $\mathcal{X}$. Non-terminal states $s \notin \cX$ correspond to ``incomplete'' objects and edges $s \to s'$ represent adding "new components" to such objects, transforming $s$ into $s'$. Let \(\mathcal{T}\) denote the set of all complete trajectories \(\tau = \left(s_0, s_1, \ldots, s_{n_{\tau}}\right)\) in the graph, where \(\tau\) is a sequence of states such that \((s_i \to s_{i + 1}) \in \mathcal{E}\) and that starts at \(s_0\) and finishes at some terminal state \(s_{n_{\tau}} \in \mathcal{X}\). As a result, any complete trajectory can be viewed as a sequence of actions that constructs the object corresponding to $s_{n_{\tau}}$ starting from the "empty object" $s_0$.

We say that a state $s'$ is a child of a state $s$ if there is an edge $(s \to s') \in \mathcal{E}$. In this case, we also say that $s$ is a parent of $s'$. Next, for any state $s$, we introduce the \textit{forward policy}, denoted by $\PF(s'|s)$ for $(s \to s') \in \cE$, as an arbitrary probability distribution over the set of children of the state $s$. In a similar fashion, we define the \textit{backward policy} as an arbitrary probability distribution over the parents of a state $s$ and denote it as $\PB(s'|s)$, where $(s' \to s) \in \cE$.

Given these two definitions, the main goal of GFlowNet training is a search for a pair of policies such that the induced distributions over complete trajectories in the forward and backward directions coincide:
\begin{equation}
\label{eq:tb}
\prod_{t=1}^{n_\tau} \PF \left(s_{t} \mid s_{t-1}\right) = \frac{\cR(s_{n_\tau})}{\rmZ} \prod_{t=1}^{n_\tau} \PB \left(s_{t-1} \mid s_{t}\right)\,, \quad \forall \tau \in \cT\,.
\end{equation}
The relation \eqref{eq:tb} is known as the \textit{trajectory balance constraint} \citep{malkin2022trajectory}. We refer to the left and right-hand sides of \eqref{eq:tb} as to the forward and backward trajectory distributions and denote them as 
\begin{equation}
\label{eq:forward_backward_traj}
\Ptraj{\PF}(\tau) := \prod_{i=1}^{n_{\tau}} \PF(s_i | s_{i-1})\,, \qquad \Ptraj{\PB}(\tau) := \frac{\cR(s_{n_{\tau}})}{\rmZ} \cdot \prod_{i=1}^{n_\tau} \PB(s_{i-1} | s_i)\eqsp,
\end{equation}
where $\tau = (s_0,s_1,\ldots,s_{n_{\tau}}) \in \cT$. If the condition \eqref{eq:tb} is satisfied for all complete trajectories, sampling a trajectory in the forward direction using \(\PF\) will result in a terminal state being sampled with probability \(\cR(x)/\mathrm{Z}\). We will call such $\PF$ a \textit{proper} GFlowNet forward policy.

In practice, we train a model (usually a neural network) that parameterizes the forward policy (and possibly other auxiliary functions) to minimize an objective function that enforces the constraint \eqref{eq:tb} or its equivalent. The main existing objectives are \textit{Detailed Balance} (\texttt{DB}, \citealp{bengio2021gflownet}), \textit{Trajectory Balance} (\texttt{TB}, \citealp{malkin2022trajectory}) and \textit{Subtrajectory Balance} (\texttt{SubTB}, \citealp{madan2023learning}). 
The \texttt{DB} objective is defined as
\begin{equation}
\label{eq:DB_loss}
\cL_{\mathrm{DB}}(\theta; s, s') = \left(\log \frac{F_\theta(s) \PF(s' | s, \theta)}{F_\theta(s') \PB(s | s', \theta)} \right)^2\,,
\end{equation}
where $F_{\theta}(s)$ is a neural network that approximates the \textit{flow} function of the state $s$, see \citep{bengio2021gflownet,madan2023learning} for more details on the flow-based formalization of the GFlowNet problem.
Using condition \eqref{eq:tb}, the \texttt{DB} objective can be generalized to entire trajectory, yielding the \texttt{TB} objective:
\begin{equation}
\label{eq:TB_loss}
\cL_{\mathrm{TB}}(\theta; \tau) = \left(\log \frac{\mathrm{Z}_\theta\prod_{t=1}^{n_{\tau}} \PF(s_t | s_{t - 1}, \theta)}{\cR(s_{n_{\tau}}) \prod_{t=1}^{n_{\tau}} \PB(s_{t - 1} | s_{t}, \theta)} \right)^2\,,
\end{equation}
where $\mathrm{Z}_\theta$ is a global constant estimating the normalizing constant $\mathrm{Z}$.
Finally, the \texttt{SubTB} objective is defined as
\begin{equation}
\label{eq:SubTB_loss}
\cL_{\mathrm{SubTB}}(\theta; \tau) = \sum\limits_{0 \le j < k \le n_{\tau}} w_{jk}\left(\log \frac{F_\theta(s_j)\prod_{t=j + 1}^{k} \PF(s_t | s_{t - 1}, \theta)}{F_\theta(s_k) \prod_{t=j + 1}^{k} \PB(s_{t - 1} | s_{t}, \theta)} \right)^2\,,
\end{equation}
here $F_{\theta}(s)$ is substituted with $\cR(s)$ for terminal states $s$, and $w_{jk}$ is usually taken to be $\lambda^{k - j}$ and then normalized to sum to $1$. \texttt{TB} and \texttt{DB} objectives can be viewed as special cases of \eqref{eq:SubTB_loss}, which are obtained by only taking the term corresponding to the full trajectory or to individual transitions, respectively. 
\par 
Some environments, such as phylogenetic tree generation, use the \texttt{FLDB} (Forward-Looking Detailed Balance) objective due to \cite{pan2023better}. This loss applies for environments with rewards $\mathcal{R} (x) = \exp(- E(x))$ with $E(x)$ is the \emph{energy function}, and the energy function is well defined for all states $s$, with the condition $E(s_0) = 0$. We then introduce the forward-looking flow function $\Tilde{F}_{\theta}(s)$, defined for a state $s$ via the flow function $F$:
\begin{equation}
\Tilde{F}_{\theta}(s) := \exp(E(s)) F_{\theta}(s)\,.
\end{equation}
Using this function, we can rewrite the detailed balance constraint and obtain the forward-looking detailed balance (FLDB) constraint (see \cite[Eq. 10]{pan2023better},). Based on this constraint, we obtain the \texttt{FLDB} objective using the same techniques as in \texttt{DB}:
\begin{equation}\label{eq:FLDB_loss}
\textstyle
\cL_{\mathrm{FLDB}}(\theta; s, s') = \left(\log \frac{\Tilde{F}_\theta(s) \PF(s' | s, \theta)}{\Tilde{F}_\theta(s') \PB(s | s', \theta)} + E(s') - E(s)\right)^2\,.
\end{equation}

All objectives allow either training the model in an on-policy regime using the trajectories sampled from $\PF$ or in an off-policy mode using the replay buffer or some exploration techniques. In addition, it is possible to either optimize $\PB$ along with $\PF$ or to use a fixed $\PB$, e.g., the uniform distribution over parents of each state. One can show that given any fixed $\PB$, there exists a unique $\PF$ that satisfies \eqref{eq:tb}; see, e.g., \citep{malkin2022trajectory}.
\par 
We further write $P_{\theta}(x)$ for a marginal distribution over the terminal set, induced by the GFlowNet with the forward policy $\PF(s' | s, \theta)$. We write $\hat P_{\theta}(x)$ for an estimate of $P_{\theta}(x)$, obtained with different environment-specific methods, detailed in further sections.

%% file: app/environment_details.tex
\section{Experiments}
\label{app:env}

In this section, we provide comparisons between \texttt{gfnx} and previous implementations of GFlowNets on various environments. We use either \texttt{torchgfn} or author implementations of the benchmarks discussed below. Reported metrics are averaged across runs on at least $3$ random seeds.

\subsection{Hypergrid}
\label{app:exp_grid}

We implement a synthetic hypergrid environment introduced by \cite{bengio2021flow}. It is a \( d \)-dimensional hypercube with a side length \( H \). The state space consists of $d$-dimensional vectors $(s_1,\ldots,s_d)^\top \in \{0,\ldots,H-1\}^d$, with the initial state being \( (0, \ldots, 0)^\top \). For each state $(s_1,\ldots,s_{d-1})$, there are at most $d+1$ actions. The first action always corresponds to an exit action that transfers the state to its terminal copy, while the remaining $d$ actions each increment one coordinate by $1$ without leaving the grid. The number of terminal states is $|\cX| = H^d$. There are $2^d$ regions with high rewards near the corners of the grid, while states outside have much lower rewards. The reward at a terminal state $s$ with coordinates $(s^1, \ldots, s^D)$ is defined as 
\begin{equation}
\cR(s) = R_0 + R_1 \times \prod_{i = 1}^D \mathbb{I}\left[0.25 < \left|\frac{s^i}{H-1}-0.5\right|\right] + R_2 \times \prod_{i = 1}^D \mathbb{I}\left[0.3 < \left|\frac{s^i}{H-1}-0.5\right| < 0.4\right]\eqsp.
\end{equation}
For our experiments we use standard reward parameters $(R_0 = 10^{-3}, R_1 = 0.5, R_2 = 2.0)$, taken from \cite{bengio2021flow}. 
\par 
This environment is sufficiently small to benchmark algorithms on a CPU and to compute the target distribution in closed form, allowing us to directly examine the convergence of empirical distribution $P_{\theta}(x)$ to $\cR(x) / \rmZ$. Moreover, this gives us access to a perfect sampler that samples from the ground truth distribution of the reward function.
\par 
We conduct experiments on a $4$-dimensional hypergrid with a side length of $20$. As an evaluation metric, we use total variation distance of the true reward distribution and the empirical distribution of the last $2 \cdot 10^5$ terminal states sampled during training. Since we compute the metric using the empirical distribution of the sampler over a finite number of samples, this introduces a bias, so even a perfect sampler does not have a zero total variation metric. Therefore, we provide the metric of a perfect sampler for this environment. We compare our implementation against the \texttt{torchgfn} library \cite{lahlou2023torchgfn} using the \texttt{DB}, \texttt{TB} and \texttt{SubTB} objectives.

\begin{figure}[!t]
    \centering
    \includegraphics[width=1\linewidth]{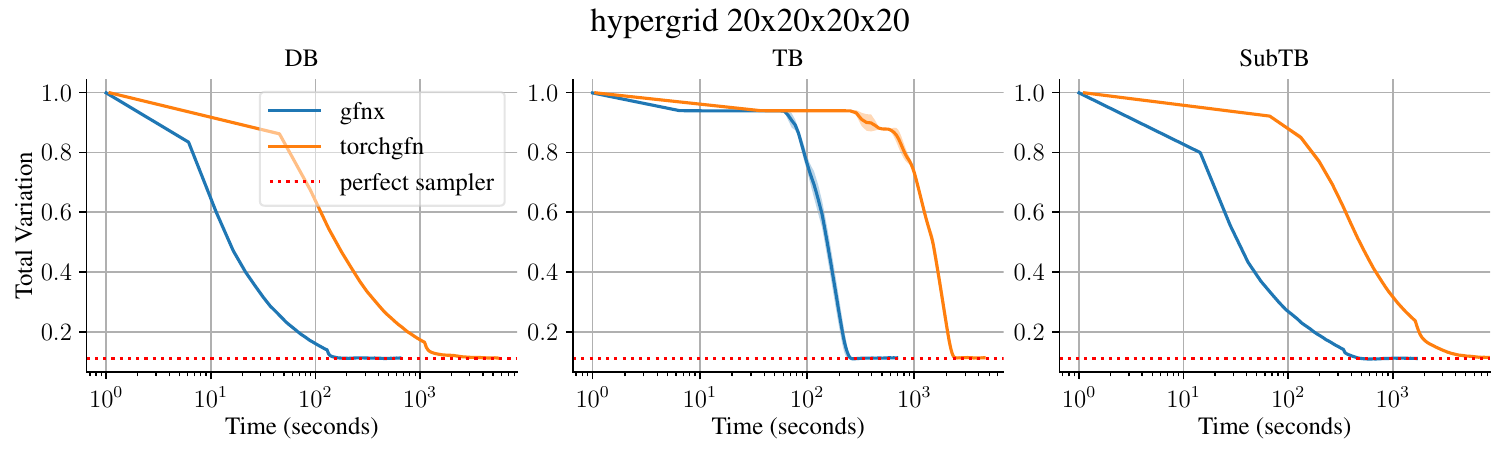}
    \caption{Total variation between true reward and empirical sample distributions versus total training time (in seconds). The same number of training iterations was used for different implementations. We use torch implementation from \cite{lahlou2023torchgfn}. These experiments were performed on CPU.} 
    \label{fig:hypergrid}
\end{figure}

\Cref{fig:hypergrid} presents the results. For all objectives, both implementations converge and yield the same total variation metric, while \texttt{gfnx} runs faster. \Cref{tab:time_exps} reports the exact performance of the libraries and shows that the \texttt{gfnx} implementation is at least about five times faster than \texttt{torchgfn} for this environment on CPU. We also provide additional experiments on $2$-dimensional hypergrid $20 \times 20$ and $8$-dimensional hypergrid with side length $10$ in \Cref{tab:hypergrid_full}.

\begin{table}[ht]
\centering
\begin{subtable}{0.48\textwidth}
\centering
\footnotesize
\begin{tabular}{@{}lcc@{}}
\toprule
Objective & \texttt{torchgfn} & \texttt{gfnx} \\
\midrule
\texttt{DB} & 321.8$_{\pm2.5}$ it/s & \textbf{3235.0$_{\pm86.7}$} it/s \\
\texttt{TB} & 359.8$_{\pm13.1}$ it/s & \textbf{1853.3$_{\pm14.9}$} it/s \\
\texttt{SubTB} & 222.1$_{\pm1.4}$ it/s & \textbf{2258.0$_{\pm53.6}$} it/s \\
\bottomrule
\end{tabular}
\caption{$2$-dimensional hypergrid with a side length of $20$.}
\label{tab:table1}
\end{subtable}\hfill
\begin{subtable}{0.48\textwidth}
\centering
\footnotesize
\begin{tabular}{@{}lcc@{}}
\toprule
Objective & \texttt{torchgfn} & \texttt{gfnx} \\
\midrule
\texttt{DB} & 209.6$_{\pm1.8}$ it/s & \textbf{1453.6$_{\pm2.9}$} it/s \\
\texttt{TB} & 225.3$_{\pm3.7}$ it/s & \textbf{1443.8$_{\pm4.0}$} it/s \\
\texttt{SubTB} & 143.6$_{\pm0.7}$ it/s & \textbf{598.1$_{\pm3.4}$} it/s \\
\bottomrule
\end{tabular}
\caption{$8$-dimensional hypergrid with a side length of $10$.}
\label{tab:table2}
\end{subtable}
\caption{Iterations per second comparison between \texttt{torchgfn} \cite{lahlou2023torchgfn} and \texttt{gfnx} libraries on small and large Hypergrid environments. Higher values indicate better performance. These experiments were performed on CPU.}
\label{tab:hypergrid_full}
\end{table}

All models are parameterized using an MLP with 2 hidden layers and 256 hidden units. We use the Adam optimizer with a learning rate of $10^{-3}$ and a batch size of 16 trajectories. For \texttt{SubTB}, we set $\lambda = 0.9$, following \cite{madan2023learning}.

Hypergrid experiments were performed on an Intel Core i7-10700F CPU with 32 GB of RAM. Table~\ref{tab:grid_params} summarizes the chosen hyperparameters. To implement the first-in first-out buffer to store the transitions in order to evaluate an empirical distribution over terminals, we used \texttt{flashbax} library \cite{flashbax}.
\input{app/table_hypergrid}

\subsection{Sequential environments}
\label{app:seq_exp}
We implement a number of sequential environments for the task of generating strings with a finite vocabulary and either fixed or variable length. Conceptually, there are several ways to model sequence generation:
\begin{itemize}
\item \textbf{Autoregressive generation (fixed length).} In this setting, sequence is generated step by step from left to right until a fixed length $n$ is reached. Each step involves selecting the next symbol from a vocabulary of size $m$. The resulting state space thus consists of all partial sequences of length up to $n$, and the action space comprises $m$ possible symbol choices at each step.
\item \textbf{Autoregressive generation (variable length).} In this setting, sequences are generated autoregressively but can terminate before reaching the maximum length $n$. To enable variable-length outputs, we introduce a special \emph{stop} action. The state space includes all strings of length up to $n$, while the action space contains the $m$ symbols plus the stop action. $\cP_B$ then, remains degenerate, as per the nature of autoregressive MDP.
\item \textbf{Prepend/append generation.}  
In this setting, a sequence of maximum length $n$ is generated by adding symbols either to the beginning or to the end of the current string. At each step, one of $m$ symbols is chosen from the vocabulary, along with a binary choice specifying whether to \emph{prepend} or \emph{append}. Thus, the action space consists of $2m$ possible actions, and the state space includes all sequences of length up to $n$.
\item \textbf{Non-autoregressive generation.}  
In this setting, we generate fixed-length sequences (of length $n$) by selecting both a position and a symbol at each step. Initial state $s_0$ corresponds to the sequence of empty tokens $\oslash$. The available actions correspond to choosing a position with an empty token and replacing it with a token from the given vocabulary. The action space can therefore be represented as all possible pairs $(i, a)$, where $i \in \{1, \dots, n\}$ denotes the position and $a \in \{1, \dots, m\}$ the symbol to insert. 
\end{itemize}

Below we present a number of popular sequential GFlowNet  environments

\paragraph{Bit sequences.} The bit sequences environment \cite{malkin2022trajectory} aims to generate binary strings of fixed length $n \in \nset$. The action space and the vocabulary size depends on the hyperparameter $k$, which is responsible for a trade-off between the trajectory length and the action space size. Formally, we choose $k | n$, splitting the bit sequence into $n / k$ blocks, each containing $k$ bits. We implement the non-autoregressive version of the environment presented in \cite{tiapkin2024generative}, where at each step one chooses a position with an empty token and replaces it with a $k$-bit word. The reward function is defined as: 
\[
\cR(x) = \exp\biggl\{-\beta \min_{x'\in M}\frac{d(x,x')}{n}\biggr\}\,,
\] 
where $d(x,x')$ is the Hamming distance between the strings $x,x'$, $M$ is the reference set (mode set), and $\beta$ is the reward exponent. We fix $M$ in advance, and the learner has access only to the reward function, not to the set $M$ itself. We generate the set $M$ according to the procedure described in \cite{malkin2022trajectory}. Namely, we choose the same size $|M| = 60$. We set
\[
H = \{'00000000', '11111111', '11110000', '00001111', '00111100'\}\eqsp.
\]
Each sequence in $M$ is generated by randomly selecting $n/8$ elements from $H$ with replacement and then concatenating them. The test set for evaluating reward correlations is generated by taking a mode and flipping $i$ random bits in it, where this is repeated for every mode and for each $0 \leq i < n$.

\begin{figure}[!ht]
    \centering
    \includegraphics[width=0.8\linewidth]{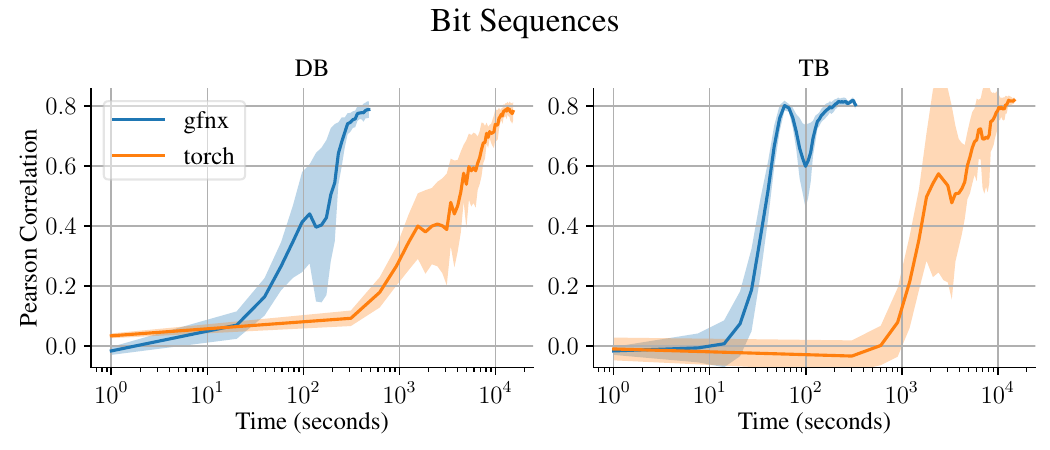}
    \caption{Comparison of \texttt{gfnx} and \texttt{torch} implementation \cite{tiapkin2024generative} on the bit sequence generation task ($n = 120, k = 8$). Comparison of the performance in terms of the Pearson correlation coefficient between the terminating state log-probability and the log-reward on $7200$ randomly sampled bit sequences (higher better). Curves were smoothed using a moving average for visual clarity. Experiments were performed on GPU using \texttt{TB} and \texttt{DB} objectives.}
    \label{fig:bitseq}
\end{figure}

We follow the evaluation setup of \cite{tiapkin2024generative}. We measure Pearson correlation on the test set between $\cR(x)$ and $\hat{P}_\theta(x)$, where the latter is a Monte Carlo estimate of the probability to sample $x$ from the trained GFlowNet proposed in \cite{zhang2022generative,tiapkin2024generative}. Namely, we notice that 
\[
P_{\theta}(x) = \mathbb{E}_{\PB(\tau \mid x)}\left[ \frac{\PF(\tau \mid \theta)}{\PB(\tau \mid x)}\right]\eqsp,
\]
and estimate $P_{\theta}(x)$ with its Monte-Carlo estimate $\hat P_{\theta}(x)$ based on $N = 10$ samples: 
\[
\hat P_{\theta}(x) = 
\frac{1}{N} \sum_{i=1}^{N} \frac{\PF(\tau^i \mid \theta)}{\PB(\tau^i \mid x)}, \quad \tau^i \sim \PB(\tau \mid x)\eqsp.
\]
Notice that any valid $\PB$ can be used here, but for each model, we take the $\PB$ that was fixed/trained alongside the corresponding $\PF$ since such a choice will lead to a lower estimate variance. 
\par 
We compare our implementation against PyTorch implementation from \cite{tiapkin2024generative} using \texttt{DB} and \texttt{TB} objectives. \Cref{fig:bitseq} presents the results. For bit sequences environment, \texttt{gfnx} implementation runs faster with the same metric values. \Cref{tab:time_exps} reports the exact performance of the implementations and shows that the \texttt{gfnx} implementation is at least $30$ times faster than \texttt{torch} on GPU.
\par 
All models are parameterized as transformers \cite{vaswani2017attention} with 3 hidden layers, 8 attention heads, and a hidden dimension of 64. Each model is trained for 50,000 iterations and a batch size of 16 with AdamW optimizer with learning rate from \( 10^{-3} \). 
\par 
To closely follow the setting of previous works~\cite{malkin2022trajectory, madan2023learning, tiapkin2024generative}, we use $\varepsilon$-uniform exploration with $\varepsilon = 10^{-3}$. Each bit sequence experiment was performed on a single NVIDIA V100 GPU. Table~\ref{tab:bitseq_tfbind8_qm9_params} summarizes the chosen hyperparameters.

\input{app/table_tfbind_qm9}

\subsubsection{TFBind8 and QM9}
\label{app:exp_tfbind_qm9}
TFBind8 environment evaluates DNA sequence design, where the goal is to generate length $8$ sequences of nucleotides. The reward is wet-lab measured DNA binding activity to a human transcription factor, SIX6 \cite{barrera2016survey}. Formally, this is an autoregressive sequence generation environment of fixed length $8$ with a vocabulary of size $4$ (corresponding to nucleotides \texttt{A}, \texttt{C}, \texttt{G}, and \texttt{T}).

QM9 is a small molecule generation environment. We use the prepend/append sequence formulation of this environment from \cite{shen2023towards}. This is a small version of the environment that uses 11 building blocks with 2 stems, and generates 5 blocks per molecule. Rewards are predictions of a proxy model trained on the QM9 dataset \cite{ramakrishnan2014qm9} to predict HOMO-LUMO gap \cite{zhang2020molecularmechanicsdrivengraphneural}. It is worth mentioning that there exists a more complex atom-by-atom setup for this environment from \cite{mohammadpour2024maximum}, while we implement a sequential formulation from \cite{shen2023towards}.

We compare our implementation against PyTorch implementation from \cite{shen2023towards} using \texttt{TB} objective. As an evaluation metric, we use total variation distance of the true reward distribution and the empirical distribution of the last $2 \cdot 10^5$ terminal states sampled during training. For both environments, we use pre-trained weights for the proxy model from \cite{shen2023towards}.

These environments are sufficiently small to allow benchmarking algorithms on a CPU and enable computation of the target distribution in closed form. Moreover, this gives us access to a perfect sampler that samples from the ground truth distribution of the reward proxy model. Since we compute the metric using the empirical distribution of the sampler, this introduces high variance, so even a perfect sampler does not have a zero total variation metric. Therefore, we provide the metric of a perfect sampler for these environments.

\begin{figure}[!t]
    \centering
    \includegraphics[width=0.8\linewidth]{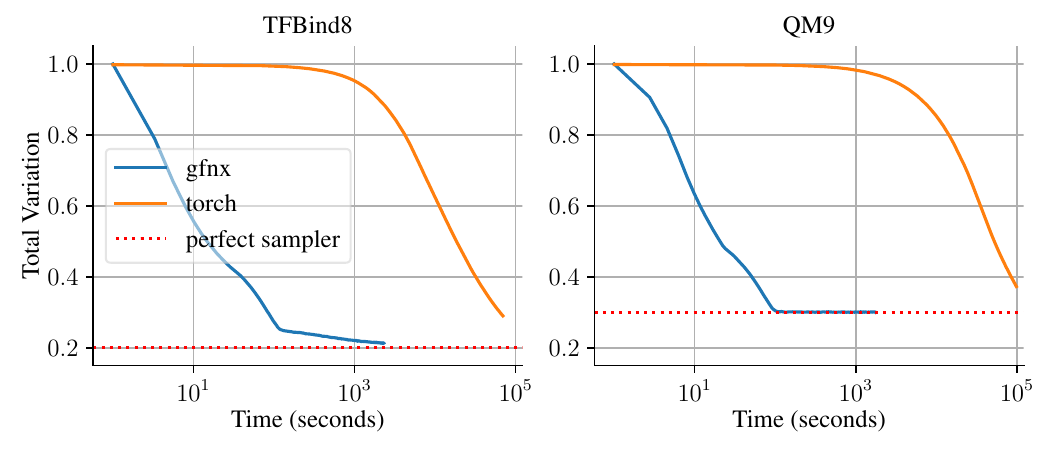}
    \caption{Total variation between true reward and empirical sample distributions versus total training time (in seconds) on TFBind8 and QM9 environments. We use torch implementation from \cite{shen2023towards}. Experiments were performed on CPU using the \texttt{TB} objective.}
    \label{fig:tfbind_qm9}
\end{figure}

\Cref{fig:tfbind_qm9} presents the results. For both TFBind8 and QM9 environments, \texttt{gfnx} implementation runs faster and more over get the better total variation metric with the same number of sampling trajectories. \Cref{tab:time_exps} reports the performance of implementations and shows that the \texttt{gfnx} implementation is at least $30$ times faster than \texttt{torch} for this environments on CPU.
\par 
All models are parameterized using an MLP with 2 hidden layers and $256$ hidden units. We use the Adam optimizer with a learning rate of $5\cdot10^{-4}$ and a batch size of $16$ trajectories. The learning rate for $\rmZ$ is fixed at $0.05$ for all experiments.
We use $\varepsilon$-uniform exploration with $\varepsilon = 1.0$ that linearly anneal to zero over the $5\cdot 10^{4}$ steps.
TFBind8 and QM9 experiments were performed on an Intel Core i7-10700F CPU with 32 GB of RAM. Table~\ref{tab:bitseq_tfbind8_qm9_params} summarizes the chosen hyperparameters.

\begin{figure}[!t]
    \centering
    \includegraphics[width=0.8\linewidth]{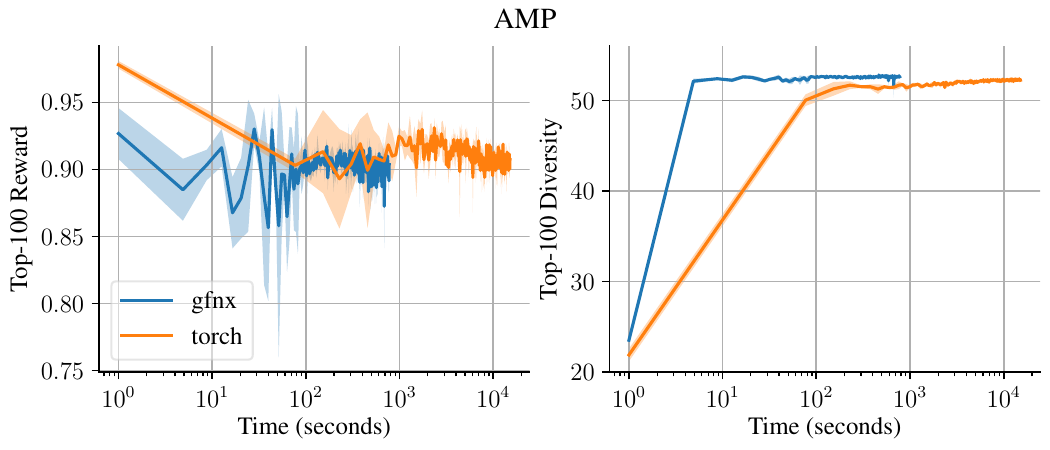}
    \caption{Top-100 reward and diversity versus total training time (in seconds) on AMP environment. We use torch implementation from \cite{jain2022biological}. The experiment was performed on GPU using the \texttt{TB} objective.}
    \label{fig:amp}
\end{figure}

\subsubsection{AMP}
\label{app:exp_amp}
In the AMP task \cite{pirtskhalava2021dbaasp}, the goal is to generate peptides with antimicrobial properties. In our setup, we use an autoregressive environment with variable sequence lengths (up to a maximum of 60) and a vocabulary consisting of 20 amino acids.

\input{app/table_amp}

We use a proxy reward model $\cR_{\varphi}(x)$ based on a classifier trained on a dataset of $3219$ AMP and $4611$ non-AMP sequences from the DBAASP database \cite{pirtskhalava2021dbaasp}. Our experimental setting is identical to the one of \cite{malkin2022trajectory}. The reward function is, therefore,
\[
\cR_{\phi}(x) = \max(\sigma(f_{\phi}(x)), r_{\min})\eqsp,
\]
where $r_{\min} > 0$ is the hyperparameter, $\sigma(\cdot)$ is a sigmoid function, and $f_{\phi}(x)$ is a classifier logit.
\par 
We also employ the evaluation setup used in (\cite{malkin2022trajectory}, \cite{jain2022biological}, \cite{madan2023learning}). We compare our implementation against PyTorch implementation from \cite{jain2022biological} using \texttt{TB} objective. \Cref{fig:amp} presents the results. For AMP environment, \texttt{gfnx} implementation runs faster with the approximately same metric values. \Cref{tab:time_exps} reports the exact performance of the implementations and shows that the \texttt{gfnx} implementation is at least $19$ times faster than \texttt{torch} for this environment on GPU.

All models are parameterized as Transformers \cite{vaswani2017attention} with 3 hidden layers, 8 attention heads, and a hidden dimension of 64. Each model is trained for 50,000 iterations and a batch size of 16 with AdamW optimizer with learning rate from \( 10^{-3} \).
We use $\varepsilon$-uniform exploration with $\varepsilon = 10^{-2}$ and initialize $\log Z = 150$ as in \cite{jain2022biological}.

For transition-level losses (e.g., \texttt{DB}), we also treat terminal and nonterminal states differently. We penalize loss highly on terminal states. We use $\lambda = 25$ as the penalization factor for terminal states. Each AMP experiment was performed on a single NVIDIA V100 GPU. Table~\ref{tab:amp_params} summarizes the chosen hyperparameters.

\subsection{Phylogenetic trees generation}
\label{app:env:phylo}
\begin{figure}[!t]
    \centering
    \includegraphics[width=1\linewidth]{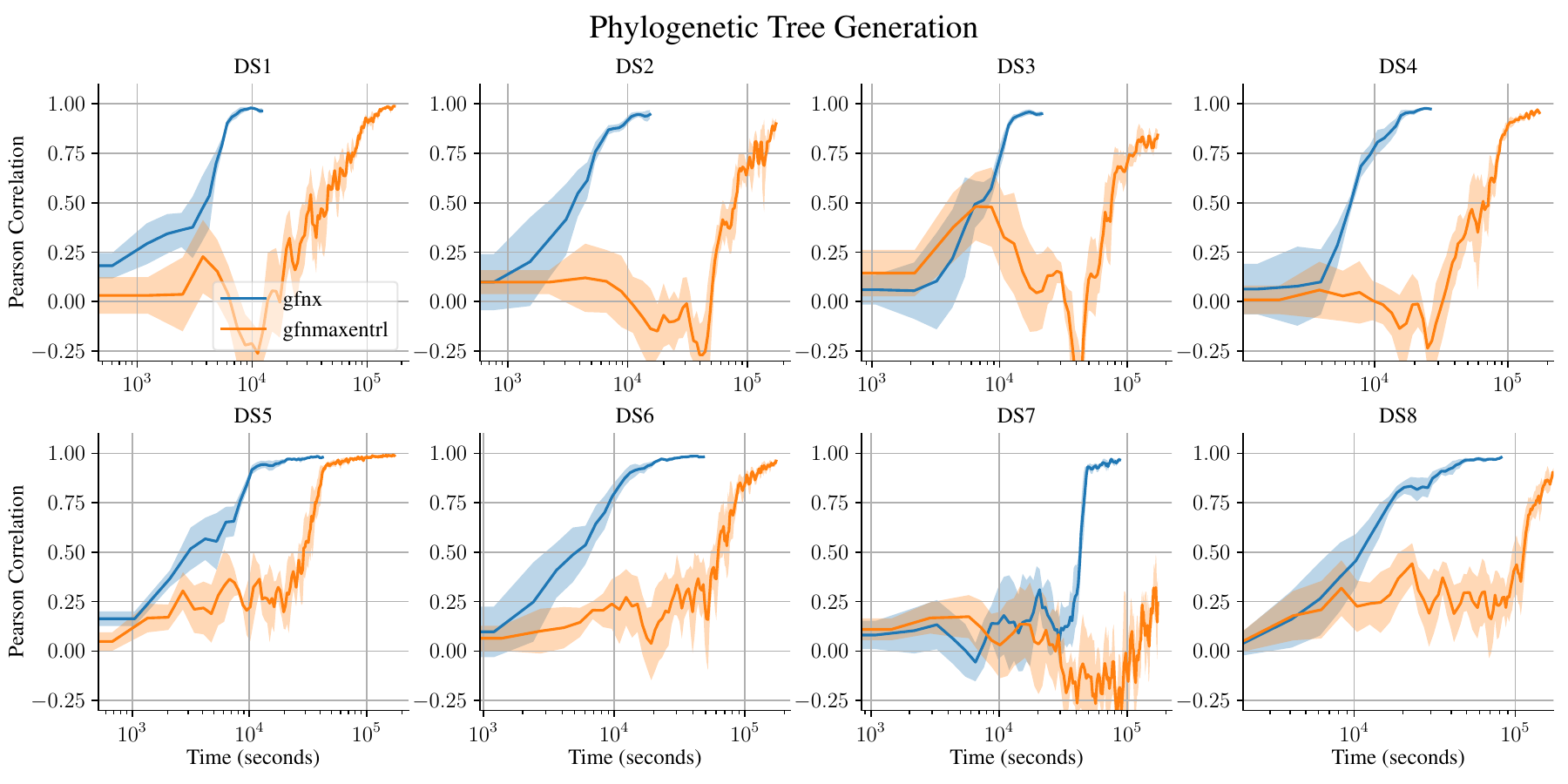}
    \caption{Comparison of \texttt{gfnx} and \cite{deleu2024discrete} paper implementation on the phylogenetic tree generation task. Comparison of the performance in terms of the Pearson correlation coefficient between the terminating state log-probability and the log-reward on $32$ randomly sampled trees (higher better). Each run is limited to $48$ hours. Curves were smoothed using a moving average for visual clarity. Experiments were performed on GPU using the \texttt{FLDB} objective.} 
    \label{fig:phylo_tree}
\end{figure}
A phylogenetic tree is a rooted binary tree whose leaves correspond to the observed species.
We follow \cite{zhou2024phylogfn} and model the construction of such trees as a sequential
decision process. The initial state $s_0$ is given by a forest of $n$ singleton trees, each containing
one species. At each step, the agent selects two trees from the forest and merges them under a new
common ancestor, resulting in a new forest with one fewer tree. After $n-1$ steps, this process
yields a complete rooted binary tree over all $n$ species. The state space therefore consists of all
possible forests that can be obtained along the way, while actions correspond to pairwise merges. The reward of a terminating state $x$ is defined as
\[
\cR(x) = \exp\left(-M(x) / \alpha\right)\,,
\]
where $\alpha > 0$ is a temperature parameter, and $M(x)$ is the parsimony score of the tree, i.e., the minimum number of mutations required
to explain the observed species under $x$. This
reward structure induces a Gibbs distribution over phylogenetic trees, strongly favoring trees with fewer evolutionary steps while still assigning non-zero probability to sub-optimal candidates. In this formulation, branch lengths are not modeled and we consider only the topology of the tree.

\input{app/table_phylo_tree}

We conduct experiments on $8$ datasets introduced in \cite{zhou2024phylogfn}. As an evaluation metric, we use the Pearson correlation coefficient between the terminating state log-probability and the log-reward of $32$ randomly sampled trees. We compare our implementation against the JAX implementation of phylogenetic tree generation from the \texttt{gfnmaxentrl} paper \cite{deleu2024discrete} using the \texttt{FLDB} objective \cite{pan2023better}. To compute correlation, we use the same Monte Carlo estimate as in the bit sequence task, we compute correlation on sampled trees from the current policy. The code in \cite{deleu2024discrete} does not provide computation of this metric, so we report self-implemented metrics.
\par 
\Cref{fig:phylo_tree} presents the results. All implementations have a compute budget of $48$ hours and were stopped after reaching it. The \cite{deleu2024discrete} implementation reached the time limit on all datasets, whereas \texttt{gfnx} works fast and finished in time. For most datasets both implementations converged to approximately the same metrics before being stopped. For the DS7 dataset, training proved particularly challenging and the \texttt{gfnmaxentrl} implementation did not converge in time. A more accurate comparison of performance is shown in \Cref{tab:time_exps}. The \texttt{gfnx} implementation is mostly at least $10$ times faster than \texttt{gfnmaxentrl}, due to the full training pipeline being written in JAX, whereas the \texttt{gfnmaxentrl} environment was written in NumPy and does not work on GPU.
\par 
For training stability we multiply the reward by a constant, depending on the dataset:
\[
    \cR(T) = \exp\left(\frac{C - M(T)}{\alpha}\right).
\]
We choose the hyperparameter $C$ as in \cite{deleu2024discrete}, for datasets DS1–DS8, we set it to $5800$, $8000$, $8800$, $3500$, $2300$, $2300$, $12500$, and $2800$, respectively. For all datasets, we use a constant reward temperature $\alpha$ set to $4$.
\par
All models are parameterized as Transformers \cite{vaswani2017attention} with $6$ hidden layers, $8$ attention heads, a hidden dimension of $32$, and a feed-forward hidden dimension of $128$. Each model is trained for $3.2 \cdot 10^6$ trajectories with the Adam optimizer. The batch size is set to $32$ for the DS1–DS4 datasets, to $16$ for the DS5, DS6, and DS8 datasets, and to $8$ for the DS7 dataset. Target models are updated using exponential moving averaging with $\tau = 0.005$.
\par 
To be more consistent with experiments for other environments, all experiments are conducted without a replay buffer, unlike the standard implementation in \cite{deleu2024discrete}. Each phylogenetic tree experiment was performed on a single NVIDIA V100 GPU. \Cref{tab:phylo_params} summarizes the chosen hyperparameters.

\subsection{Structure learning of Bayesian networks}
\label{env:dag}
We consider the task of Bayesian structure learning. A Bayesian network is a probabilistic graphical model over random variables $\{X_{1}, \ldots, X_{d}\}$, where the joint distribution factorizes according to a DAG $G$ as
\begin{equation}
    P(X_{1}, \ldots, X_{d}) = \prod_{k=1}^{d} P\big(X_{k} \mid \mathrm{Pa}_{G}(X_{k})\big)\,,
\end{equation}
with $\mathrm{Pa}_{G}(X_{k})$ the parent set of $X_{k}$ in $G$. The goal of Bayesian structure learning is to approximate the posterior
\begin{equation}
    P(G \mid \mathcal{D}) \propto P(\mathcal{D} \mid G) P(G)\,,
\end{equation}
where $\mathcal{D}$ denotes the observed dataset. We assume that the samples in $\mathcal{D}$ are i.i.d., fully-observed. We use the setting where $\mathcal{D}$ is generated under Erd\H{o}s-R\'{e}nyi model. We also assume a uniform prior over structures and implement both the \textbf{B}ayesian metric for \textbf{G}aussian networks having score \textbf{e}quivalence (BGe) \citep{geiger1994learning} and 
linear Gaussian \citep{nishikawa2022bayesian}
scores as marginal likelihoods $P(\mathcal{D} \mid G)$.

The environment is defined as a sequential process that constructs a DAG by adding edges one at a time, while enforcing acyclicity by maintaining an incrementally updated adjacency matrix and transitive closure similar to \citep{deleu2022bayesian}. A special stop action yields a terminal state corresponding to a valid DAG with a reward term that is given by
\begin{equation}
    \log \cR(G) = \log P(\mathcal{D} \mid G) + \log P(G)\,.
\end{equation}
Since in the environment every state can be terminal, we can employ Modified Detailed Balance (MDB) \citep{deleu2022bayesian} algorithm for learning the posterior distribution. We also take the advantage of the reward structure for an efficient computation of the delta score, required by the MDB loss. 

Experiments are conducted on $20$ random Erd\H{o}s--R\'enyi ground-truth graphs with expected in-degree $1$, where for each graph we generate $100$ samples by ancestral sampling. This problem is typically studied with $d=5$ nodes, although settings with $d=20$ or more are also possible. This environment is benchmarked only on GPU, as our models include deep architectures such as transformers and graph neural networks. We evaluate methods using Jensen--Shannon divergence between the learned and exact distributions over sampled DAGs. Additionally we implement correlation scores over path, edge, and Markov blanket marginals.

\begin{figure}[t]
    \centering
    \includegraphics[width=0.4\linewidth]{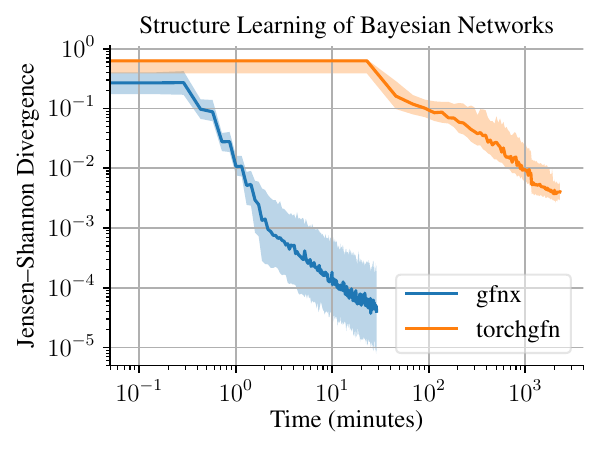}
    \caption{Jensen–Shannon divergence between true reward and empirical sample distributions versus total training time (in minutes). We use torch implementation from \cite{lahlou2023torchgfn}. The experiment was performed on GPU using the \texttt{MDB} objective.}
    \label{fig:structure_learning}
\end{figure}

\paragraph{Local Scores and Modularity.}
\label{app:env:dag:modularity}
A fundamental property of Bayesian structure learning is the \emph{modularity} of both the structure prior $P(G)$ and the parameter prior $P(\phi \mid G)$ \citep{heckerman1995bde,chickering1995learning}. Under this assumption, the score of a graph decomposes into independent contributions from each variable given its parent set. Specifically, for a Bayesian network with DAG $G$ over $d$ nodes, the log-reward admits the additive form
\begin{equation}
    \log \cR(G) \;=\; \sum_{j=1}^{d} \mathrm{LocalScore}\big(X_{j} \mid \mathrm{Pa}_{G}(X_{j})\big)\,,
\end{equation}
where each term $\mathrm{LocalScore}(X_{j} \mid \mathrm{Pa}_{G}(X_{j}))$ depends only on $X_{j}$ and its parents. Standard Bayesian scores such as Linear Gaussian and BGe satisfy this decomposition.

Modularity implies that reward updates are computationally efficient. When an edge $X_{i} \to X_{j}$ is added to the graph, all local scores remain unchanged except for that of $X_{j}$. Consequently, the change in log-reward reduces to
\begin{align}
    \log \cR(G') - \log \cR(G) 
    &= \mathrm{LocalScore}\big(X_{j} \mid \mathrm{Pa}_{G}(X_{j}) \cup \{X_{i}\}\big)  - \mathrm{LocalScore}\big(X_{j} \mid \mathrm{Pa}_{G}(X_{j})\big)\,,
\end{align}
a quantity referred to as the \emph{delta score} or \emph{incremental value} \citep{friedman2003ordermcmc}. The delta score requires updating only a single local component, making it efficient to compute. This property is widely used in structure learning search algorithms \citep{chickering2002ges,koller2009pgm}, and in our setting enables efficient computation of rewards within the GFlowNet training objective.

\paragraph{Dataset Generation Process.}
\label{app:env:dag:dataset}
For our experiments, datasets are generated from randomly sampled ground-truth Bayesian networks. Each ground-truth DAG $G^{*}$ is sampled from an Erd\H{o}s--R\'enyi distribution with $d=5$ nodes and an expected in-degree of $1$. Conditional probability distributions are assumed to be linear-Gaussian: for each variable $X_{j}$, given its parents $\mathrm{Pa}_{G^{*}}(X_{j})$, we define
\begin{equation}
    X_{j} \mid \mathrm{Pa}_{G^{*}}(X_{j}) \sim \mathcal{N}\Bigg(\sum_{i \in \mathrm{Pa}_{G^{*}}(X_{j})} w_{ij} X_{i}, \; \sigma^{2}_{j}\Bigg)\,,
\end{equation}
where the weights $w_{ij}$ are drawn according to $w_{ij} \sim \mathcal{N}(0, 1)$ and the noise variances are fixed to $\sigma^{2}_{j} = 0.1$. Data $\mathcal{D}$ is then generated by ancestral sampling: nodes are ordered topologically according to $G^{*}$, and each variable is sampled conditionally on its parents. For each DAG, we generate $100$ observations in this way. 

This procedure is repeated for $20$ different random seeds, producing a collection of datasets associated with $20$ distinct Erd\H{o}s--R\'enyi graphs. Each dataset therefore consists of $100$ samples from a linear-Gaussian Bayesian network with known ground-truth structure, which enables exact posterior computation and comparison against learned distributions.

\input{app/table_structure_learning}

\paragraph{Online Mask Updates.}
\label{app:env:dag:mask}
To enforce acyclicity efficiently, we maintain two structures for each DAG $G$: the adjacency matrix of $G$, and the adjacency matrix of the transitive closure of $G^{\top}$ (the transpose graph obtained by reversing all edge directions). The mask of legal actions is derived from these components: it excludes (i) edges already present in $G$, and (ii) edges whose addition would create a cycle according to the transitive closure. When a new edge $(u \to v)$ is added, the update is performed online by (1) setting the corresponding entry in the adjacency matrix, and (2) updating the transitive closure of $G^{\top}$ via the outer product of the column of $v$ and the row of $u$, applied as a binary OR to the existing closure. This update rule ensures correctness and requires only $O(d^{2})$ time per edge addition, avoiding expensive cycle checks at every step.

\paragraph{Evaluation Metrics and Structural Features.}
\label{app:env:dag:stats}

We compare the learned distribution over DAGs with the exact posterior using the Jensen--Shannon divergence (JSD). We also provide an implementation of structural feature marginals correlations. Since the number of DAGs with $d=5$ nodes is finite ($29{,}281$), all probabilities can be computed exactly by enumeration.

\Cref{fig:structure_learning} presents the results. All runs were given a $48$-hour compute budget. The \texttt{torchgfn} baseline was trained with its replay-buffer implementation (the only stable variant), while \texttt{gfnx} was evaluated in a fully on-policy setting, which is substantially more computationally demanding. Despite this disadvantage, \texttt{gfnx} converges faster and reaches lower JSD, outperforming \texttt{torchgfn} by roughly a factor of $79\times$ in wall-clock efficiency under the BGe score. Both methods follow comparable learning trajectories, but \texttt{torchgfn} approaches the time limit well before matching the threshold score achieved by \texttt{gfnx}. \Cref{tab:dag_params} summarizes the chosen hyperparameters.

\subparagraph{Jensen--Shannon divergence.}
For two distributions $P$ and $Q$ over the same support, the JSD is defined as
\begin{equation}
    \mathrm{JSD}(P \parallel Q) = \tfrac{1}{2}\,\mathrm{KL}(P \parallel M) + \tfrac{1}{2}\,\mathrm{KL}(Q \parallel M)\,,
\end{equation}
where $M = \tfrac{1}{2}(P+Q)$ and $\mathrm{KL}$ denotes the Kullback--Leibler divergence.

\subparagraph{Edge features.}
For nodes $X_{i}, X_{j}$, the marginal probability of the edge feature $X_{i}\to X_{j}$ is
\begin{equation}
    P(X_{i}\to X_{j}\mid \mathcal{D}) = \sum_{G} \mathbf{1}(X_{i}\to X_{j}\in G)\, P(G\mid \mathcal{D})\,.
\end{equation}

\subparagraph{Path features.}
For nodes $X_{i}, X_{j}$, the path feature $X_{i}\rightsquigarrow X_{j}$ indicates the existence of a directed path from $X_{i}$ to $X_{j}$. Its marginal probability is
\begin{equation}
    P(X_{i}\rightsquigarrow X_{j}\mid \mathcal{D}) = \sum_{G} \mathbf{1}(X_{i}\rightsquigarrow X_{j}\in G)\, P(G\mid \mathcal{D})\,.
\end{equation}

\subparagraph{Markov blanket features.}
For nodes $X_{i}, X_{j}$, the Markov blanket feature $X_{i}\sim_{M} X_{j}$ indicates that $X_{i}$ belongs to the Markov blanket of $X_{j}$, i.e.\ $X_{i}$ is either a parent, a child, or a co-parent of $X_{j}$. Its marginal probability is
\begin{equation}
    P(X_{i}\sim_{M} X_{j}\mid \mathcal{D}) = \sum_{G} \mathbf{1}(X_{i}\sim_{M} X_{j}\in G)\, P(G\mid \mathcal{D})\,.
\end{equation}


%% file: app/table_hypergrid.tex
\begin{table}[t!]
\begin{center}
\caption{Hyperparameter choices for hypergrid experiments.}
\label{tab:grid_params}
\vspace{0.3cm}
\begin{tabular}{|l|c|}
\hline
Hyperparatemeter & Value \\
\hline \hline
    Training trajectories & $10^6$ \\ 
    Learning rate & $10^{-3}$ \\
    $Z$ learning rate for \TB & $10^{-1}$  \\
    Adam optimizer $\beta, \epsilon, \lambda$ & $(0.9, 0.999), 10^{-8}, 0$ \\ 
    Number of hidden layers & $2$ \\
    Hidden embedding size & $256$ \\
\hline
    \SubTB $\lambda$ & $0.9$ \\
\hline
\end{tabular}
\end{center}
\end{table}

%% file: app/table_tfbind_qm9.tex
\begin{table}[t!]
\begin{center}
\caption{Hyperparameter choices for Bit Sequences, TFBind8, and QM9 experiments.}
\label{tab:bitseq_tfbind8_qm9_params}
\vspace{0.3cm}
\begin{tabular}{|l|lr|}
\hline Hyperparatemeter & Bit Sequences & TFBind8 and QM9  \\
\hline \hline
Training iterations & $5 \times 10^4$ & $10^6$ \\ 
Learning rate & $10^{-3}$ & $5 \times 10^{-4}$ \\
$Z$ learning rate for \TB & \multicolumn{2}{c|}{$0.05$}  \\
Adam optimizer $\beta, \epsilon, \lambda$ & 
$(0.9, 0.999), 10^{-8}, 10^{-5}$ & $(0.9, 0.999), 10^{-8}, 0$ \\ 
$\varepsilon$-uniform exploration & $10^{-3}$ & $1.0$ linearly annealed to $0.0$ \\
Reward exponent & $3$ & $10$ \\ 
Batch size & \multicolumn{2}{c|}{$16$} \\
Number of MLP layers &  -- & $2$ \\
Number of transformer layers & $3$ & -- \\
Hidden embedding size & $64$ & $256$ \\
Dropout & 0 & -- \\
\hline
\end{tabular}
\end{center}
\end{table}

%% file: app/table_amp.tex
\begin{table}[ht!]
\begin{center}
\caption{Hyperparameter choices for AMP experiment.}
\label{tab:amp_params}
\vspace{0.3cm}
\begin{tabular}{|l|c|}
\hline
Hyperparatemeter & Value \\
\hline\hline 
    Training iterations & $2 \times 10^4$ \\ 
    Learning rate & $10^{-3}$ \\
    $Z$ learning rate for \TB & $0.64$  \\
    Adam optimizer $\beta, \epsilon, \lambda$ & $(0.9, 0.999), 10^{-8}, 10^{-5}$ \\ 
    $\varepsilon$-uniform exploration & $10^{-3}$ \\
    Number of transformer layers & $3$ \\
    Hidden embedding size & $64$ \\
    Dropout & $0.0$ \\
\hline
\end{tabular}
\end{center}
\end{table}

%% file: app/table_phylo_tree.tex
\begin{table}[t!]
\begin{center}
\caption{Hyperparameter choices for phylogenetic trees generation experiments.}
\label{tab:phylo_params}
\vspace{0.3cm}
\begin{tabular}{|l|c|}
\hline Hyperparatemeter & Value  \\
\hline \hline
Reward temperature $\alpha$ & $4$ \\
Reward constant $C$ & \{5800, 8000, 8800, 3500, 2300, 2300, 12500, 2800\} \\
Training trajectories & $3.2\times 10^6$ \\ 
Learning rate & 
$3 \cdot 10^{-4}$ cosinely annealed to $10^{-5}$ \\
Warmup steps & $5000$ \\
Adam optimizer $\beta, \epsilon, \lambda$ & 
$(0.9, 0.999), 10^{-8}, 0$ \\ 
$\varepsilon$-uniform exploration & $1.0$ linearly annealed to $0.0$ for half of training \\
\hline
Number of attention heads & $8$ \\
Number of transformer layers & $6$ \\
Hidden size & $32$ \\
Dropout & $0.01$ \\
Embedding size & $128$ \\
Hidden embedding size & $256$ \\
Number of MLP layers & $3$ \\
MLP hidden size & $256$ \\
\hline
\end{tabular}
\end{center}
\end{table}

%% file: app/table_structure_learning.tex
\begin{table}[t!]
\begin{center}
\caption{Hyperparameter choices for structure learning of Bayesian networks experiments.}
\label{tab:dag_params}
\vspace{0.3cm}
\begin{tabular}{|l|c|}
\hline Hyperparatemeter & Value  \\
\hline \hline
Training steps & $100000$ \\
Batch size & $128$ \\
Learning rate & $10^{-4}$ \\
$\varepsilon$-uniform exploration & $1.0$ linearly annealed to $0.1$ for half of training \\
Target network update frequency & 1000 \\
Number of GNN layers & 1 \\
Number of MLP layers & 2 \\
Embedding size & 128 \\
Hidden size & 128 \\
Number of transformer heads & 4 \\
\hline
\end{tabular}
\end{center}
\end{table}

%% file: app/experimental_details.tex
\subsection{Ising model}
\label{app:exp::ising_model}
We consider the problem of jointly learning an energy-based reward model (EBM) and a generative policy, given only samples stored in a dataset. An example of such a setting is learning the interaction matrix of an Ising model \citep{ising1925beitrag} together with a corresponding GFlowNet sampler, as explored in \citep{zhang2022generative}. The proposed algorithm alternates updates between a Generative Flow Network and an energy model. A trajectory is generated either from the current forward policy, $\tau \sim P_F(\tau)$, with probability $\alpha$ (a hyperparameter of the procedure), or by first sampling a random data point $\mathbf{x}_i$ from the dataset and then generating a full trajectory using the backward policy, $\tau \sim P_B(\tau|\mathbf{x}_i)$, with probability $1 - \alpha$. The GFlowNet is then updated according to Equation~\ref{eq:TB_loss}, using the current approximation of the energy-based reward model, $R(\mathbf{x}) = \exp(-\mathcal{E}(\mathbf{x}; \varphi))$. 

The second step of the algorithm updates the parameters of the EBM. A common approach employs the contrastive divergence loss \citep{hinton2002training}, which yields the following stochastic gradient:
\begin{equation}
    \mathbb{E}_{\mathbf{x} \sim p_{\text{data}}(\mathbf{x})}
    \big[\nabla_{\varphi} \mathcal{E}_{\varphi}(\mathbf{x})
    - 
    \mathbb{E}_{\mathbf{x'} \sim q_{K}(\mathbf{x'|\mathbf{x}})}
    \big[\nabla_{\varphi} \mathcal{E}_{\varphi}(\mathbf{x'})\big]\big]\,.
\end{equation}

In the context of Generative Flow Networks, the conditional distribution in the second term, $q_{K}(\mathbf{x'}|\mathbf{x})$, is obtained by performing $K$ backward steps from the true sample $\mathbf{x}$ using the backward policy $P_B(\cdot|\mathbf{x})$, followed by $K$ forward steps using the forward policy $P_F$. The resulting sample $\mathbf{x}'$ is then either accepted or rejected according to the acceptance probability:
\begin{equation}
A (\mathbf{x} \rightarrow \mathbf{x}') = 
\min \left[ 
1, 
\frac{e^{-\mathcal{E}_{\varphi}(\mathbf{x'})}}{e^{-\mathcal{E}_{\varphi}(\mathbf{x})}} 
\frac{P_B(\tau|\mathbf{x}) \, P_F(\tau')}{P_B(\tau'|\mathbf{x'}) \, P_F(\tau)} 
\right]\,.
\end{equation}
Further details on the algorithm can be found in \citep{zhang2022generative}.

In our framework, we consider an Ising model defined on a given interaction graph with $D=N^2$ spins $\mathbf{x}^i \in \{-1, +1\}$ and energy function 
\begin{equation}
\mathcal{E}_{J}(\mathbf{x}) = -\mathbf{x}^\top J \mathbf{x}, \ J \in \mathbb{R}^{D \times D}\eqsp.
\end{equation}
This defines a probability distribution over lattice configurations:
\begin{equation}
    P(\mathbf{x}) \propto \exp(-\mathcal{E}_{J}(\mathbf{x})), \quad \mathbf{x} \in \{-1, +1\}^D\,.
\end{equation}

\begin{table}
\vspace{-4mm}
    \centering
    \resizebox{\linewidth}{!}{
   \begin{tabular}{lccccccc}
\toprule
 & \multicolumn{5}{c}{$D=10^2$} & \multicolumn{2}{c}{$D=9^2$} \\
\cmidrule(lr){2-6}\cmidrule(lr){7-8}
 Method $\backslash$ $\sigma$ & $0.1$ & $0.2$ & $0.3$ & $0.4$ & $0.5$ & $-0.1$ & $-0.2$  \\ 
 \midrule
EB-GFN (gfnx) & $4.93 \pm 0.25$ & $4.48 \pm 0.3$ & $4.05 \pm 0.03$ & $3.26 \pm 0.06$ & $2.72 \pm 0.04$  & $3.79 \pm 0.15$ & $3.98 \pm 0.01$ \\
\bottomrule
    \end{tabular}
    }
    \caption{Mean negative log-RMSE (higher is better) between data-generating matrix $J$ and learned matrix $J_{\varphi}$ for different values of $\sigma$.}
    \label{tab:ising_results}
\vspace{-5mm}
\label{tab:ising_model}
\end{table}

Following \citep{zhang2022generative}, we model sampling from the corresponding Gibbs distribution as a GFlowNet environment over partial assignments. States are represented as ternary vectors $s \in \{-1, +1, \varnothing\}^D$, with the initial state $s_0 = (\varnothing, \dots, \varnothing)$. At each step, an action selects an unassigned site and sets its spin to $-1$ or $+1$; the backward policy, conversely, removes a previously assigned spin. After $D$ steps, the process terminates at a full configuration $\mathbf{x} \in \{-1, +1\}^D$.

We define the true interaction matrix as $J = \sigma A_N$, where $\sigma \in \mathbb{R}$ and $A_N$ is the adjacency matrix of a toroidal lattice with side length $N$. Experiments are conducted on lattices with side $N = 9$ and negative couplings $\sigma \in \{-0.1, -0.2\}$, and with $N = 10$ and positive couplings $\sigma \in \{0.1, 0.2, 0.3, 0.4, 0.5\}$. We also set $K = D$, so that $q_{K}(\mathbf{x'}|\mathbf{x}) = P_T(\mathbf{x'})$, where $P_T(\mathbf{x'})$ denotes the marginal probability of sampling $\mathbf{x'}$ from the GFlowNet.

\Cref{tab:ising_model} reports the mean negative log-RMSE between the learned interaction matrix and the true lattice couplings for various values of~$\sigma$. This is the only available open-source implementation of the EB-GFN setting from \cite{zhang2022generative}. The scores align with the baselines reported in \cite{zhang2022generative}, though they differ from the exact numbers of their original implementation. Performance remains stable across lattice sizes and coupling strengths, with higher accuracy for weaker interactions.

To generate the dataset of true samples, we employ MCMC-based methods, including the Wolff algorithm \cite{wang1990cluster} and Heat Bath Parallel Tempering \cite{hukushima1996exchange}. \Cref{tab:ising_params} summarizes the chosen hyperparameters. We follow the same hyperparameter search procedure, introduced in \citep{zhang2022generative}. Note that, according to their work, the training procedure stops when the error between the trained $J_{\varphi}$ and the true $J$ reaches its minimum. 

\input{app/table_ising}

%% file: app/table_ising.tex
\begin{table}[t!]
\begin{center}
\caption{Hyperparameter choices for Ising model experiments.}
\label{tab:ising_params}
\vspace{0.3cm}
\begin{tabular}{|l|c|}
\hline Hyperparatemeter & Value  \\
\hline \hline
Training steps & $20000$ \\
Batch size & $256$ \\
MLP hidden size & 256 \\
MLP depth & 4 \\
Number of true data samples & 2000 \\
\hline
\end{tabular}
\end{center}
\end{table}